\documentclass{article} % For LaTeX2e
\usepackage{nips15submit_e,times}
\usepackage{hyperref}
\usepackage{url}
\usepackage{graphicx}
\usepackage{amsmath}
\usepackage{amssymb}
\usepackage{bm}
\usepackage{wrapfig}

\title{Stochastic Expectation Propagation for Large Scale Gaussian Process Classification}

\author{
Daniel Hern\'andez-Lobato\\
Universidad Aut\'onoma de Madrid\\
\texttt{daniel.hernandez@uam.es} 
\And
Jos\'e Miguel Hern\'andez-Lobato \\
Harvard University \\
\texttt{jmhl@seas.harvard.edu} \\
\And
Yingzhen Li\\
University of Cambridge\\
\texttt{yl494@cam.ac.uk}\\
\And
Thang Bui\\
University of Cambridge\\
\texttt{tdb40@cam.ac.uk}\\
\And
Richard E. Turner\\
University of Cambridge\\
\texttt{ret26@cam.ac.uk}
}

\nipsfinalcopy % Uncomment for camera-ready version

\begin{document}

\maketitle

\begin{abstract}
\vspace{-.1cm}
A method for large scale Gaussian process classification has been recently proposed based on 
expectation propagation (EP). Such a method allows Gaussian process classifiers to be trained 
on very large datasets that were out of the reach of previous deployments of EP and has been 
shown to be competitive with related techniques based on stochastic variational inference. 
Nevertheless, the memory resources required scale linearly with the dataset size, 
unlike in variational methods. This is a severe limitation when the number of instances 
is very large.  Here we show that this problem is avoided when stochastic EP is used to train the model.
\end{abstract}

\vspace{-.2cm}

\section{Introduction}

\vspace{-.1cm}

Gaussian process classifiers are a very effective family of non-parametric methods for 
supervised classification \cite{rasmussen2005book}. In the binary case, the class label $y_i \in \{-1,1 \}$ 
associated to each data instance $\mathbf{x}_i$ is assumed to depend on the sign of a 
function $f$ which is modeled using a Gaussian process prior. Given some data $\mathcal{D}=\{(\mathbf{x}_i,y_i)\}_{i=1}^n$, 
learning is performed by computing a posterior distribution for $f$. Nevertheless,  
the computation of such a posterior distribution is intractable and it must be approximated 
using methods for approximate inference \cite{nickish2008}. A practical disadvantage is that the cost 
of most of these methods scales like $\mathcal{O}(n^3)$, where $n$ is the number of training instances. 
This limits the applicability of Gaussian process classifiers to small datasets with a few data instances at most. 

Recent advances on Gaussian process classification have led to sparse methods of approximate 
inference that reduce the training cost of these classifiers. Sparse methods introduce 
$m \ll n$ inducing points or pseudoinputs, 
whose location is determined during the training process, leading to a training cost 
that is $\mathcal{O}(m^2n)$ \cite{quinonero2005,Snelson2006,NIPS2007_552}. A notable approach 
combines in \cite{HensmanMG15} the sparse approximation suggested in \cite{Titsias-09} with stochastic 
variational inference  \cite{hoffman13a}. This allows to learn the posterior for $f$ 
and the hyper-parameters (inducing points, length-scales, amplitudes and noise) using 
stochastic gradient ascent. The consequence is that the training cost is $\mathcal{O}(m^3)$, which does 
not depend on the number of instances $n$. Similarly, in a recent work, expectation propagation (EP) \cite{minka2001}
is considered as an alternative to stochastic variational inference for training these classifiers 
\cite{herandezlobato2015}. That work shows (i) that stochastic gradients can also be used to learn the 
hyper-parameters in EP, and (ii) that EP performs similarly to the variational approach, but does not 
require one-dimensional quadratures.

A disadvantage of the approach described in  \cite{herandezlobato2015} is that the memory 
requirements scale like $\mathcal{O}(nm)$ since EP stores in memory $\mathcal{O}(m)$ parameters 
for each data instance. This is a severe limitation when dealing with very large 
datasets with millions of instances and complex models with many inducing points. To reduce 
the memory cost, we investigate in this extended abstract, as an alternative to EP,
the use of stochastic propagation (SEP) \cite{Yingzhen2015}. Unlike EP, SEP only stores a single global approximate 
factor for the complete likelihood of the model, leading to a memory cost that scales like $\mathcal{O}(m^2)$.

\section{Large scale Gaussian process classification via expectation propagation}

\vspace{-.2cm}

We now explain the method for Gaussian process classification described in 
\cite{herandezlobato2015}. Consider $\mathbf{y}=(y_1,\ldots,y_n)$ the observed 
labels. Let $\mathbf{X}=(\mathbf{x}_1,\ldots,\mathbf{x}_n)^\text{T}$ be a matrix with the
observed data. The assumed labeling rule is $y_i = \text{sign}(f(\mathbf{x}_i) +\epsilon_i)$, where $f(\cdot)$ 
is a non-linear function following a zero mean Gaussian process with covariance function 
$k(\cdot,\cdot)$, and $\epsilon_i$ is standard normal noise that accounts for mislabeled data.  
Let $\overline{\mathbf{X}} = (\overline{\mathbf{x}}_1,\ldots,\overline{\mathbf{x}}_m)^\text{T}$ be 
the matrix of inducing points (\emph{i.e.}, virtual data that specify how $f$ varies).
Let $\mathbf{f}=(f(\mathbf{x}_1),\ldots,f(\mathbf{x}_n))^\text{T}$ and 
$\overline{\mathbf{f}}=(f(\overline{\mathbf{x}}_1),\ldots,f(\overline{\mathbf{x}}_n))^\text{T}$ be the vectors of
$f$ values associated to $\mathbf{X}$ and $\overline{\mathbf{X}}$, respectively. 
The posterior of $\mathbf{f}$ is approximated as $p(\mathbf{f}|\mathbf{y}) \approx
p(\mathbf{f}|\overline{\mathbf{f}})q(\overline{\mathbf{f}})d\overline{\mathbf{f}}$, with $q$ a Gaussian 
that approximates $p(\overline{\mathbf{f}}|\mathbf{y})$, \emph{i.e.}, the posterior of the values
associated to $\overline{\mathbf{X}}$. To get $q$, first the full independent 
training conditional approximation (FITC) \cite{quinonero2005} of $p(\mathbf{f}|\overline{\mathbf{f}})$
is employed to approximate $p(\overline{\mathbf{f}}|\mathbf{y})$ and to reduce
the training cost from $\mathcal{O}(n^3)$ to $\mathcal{O}(m^2n)$:
\vspace{-.05cm}
\begin{align}
p(\overline{\mathbf{f}}|\mathbf{y}) &=  \frac{\int p(\mathbf{y}|\mathbf{f}) p(\mathbf{f}|\overline{\mathbf{f}}) 
	d \mathbf{f}p(\overline{\mathbf{f}}|\overline{\mathbf{X}}) }{p(\mathbf{y}|\overline{\mathbf{X}})} \approx 
\frac{\int p(\mathbf{y}|\mathbf{f}) p_\text{FITC}(\mathbf{f}|\overline{\mathbf{f}})
	d \mathbf{f}p(\overline{\mathbf{f}}|\overline{\mathbf{X}}) }{p(\mathbf{y}|\overline{\mathbf{X}})} 
=\frac{ \prod_{i=1}^n \phi_i(\overline{\mathbf{f}}) p(\overline{\mathbf{f}}|\overline{\mathbf{X}}) }{p(\mathbf{y}|\overline{\mathbf{X}})} 
\,, 
\label{eq:posterior} 
\end{align} \par
\vspace{-.3cm}
where $p(\mathbf{y}|\mathbf{f})=\prod_{i=1}^n \Phi(y_i f_i)$, 
$p_\text{FITC}(\mathbf{f}|\overline{\mathbf{f}})=\prod_{i=1}^n 
p(f_i|\overline{\mathbf{f}})=\prod_{i=1}^n \mathcal{N}(f_i|m_i,s_i)$
and $\phi_i(\overline{\mathbf{f}}) = \int \Phi(y_i f_i) 
\allowbreak
\mathcal{N}(f_i|m_i ,s_i )d f_i = \Phi(y_i m_i / \sqrt{s_i + 1})$, with 
$m_i = \mathbf{K}_{f_i\overline{\mathbf{f}}}\mathbf{K}_{\overline{\mathbf{f}}\overline{\mathbf{f}}}^{-1} 
\overline{\mathbf{f}}$, $s_i = \mathbf{K}_{f_if_i} - 
\mathbf{K}_{f_i\overline{\mathbf{f}}}\mathbf{K}_{\overline{\mathbf{f}}\overline{\mathbf{f}}}^{-1} 
\mathbf{K}_{\overline{\mathbf{f}}f_i}$,
$p(\overline{\mathbf{f}}|\overline{\mathbf{X}})=\mathcal{N}(\overline{\mathbf{f}}|\mathbf{0},
\mathbf{K}_{\overline{\mathbf{f}}\overline{\mathbf{f}}})$ and $p(\mathbf{y}|\overline{\mathbf{X}})$ 
is the marginal likelihood. Furthermore, $\mathbf{K}_{\overline{\mathbf{f}} \overline{\mathbf{f}}}$ is
a matrix with the prior covariances among the entries in $\overline{\mathbf{f}}$, 
$\mathbf{K}_{f_i \overline{\mathbf{f}}}$ is a row vector with the prior covariances 
between $f_i$ and $\overline{\mathbf{f}}$ and $\mathbf{K}_{f_if_i}$ is the prior variance of $f_i$.
Finally, $\mathcal{N}(\cdot|\mathbf{m},\bm{\Sigma})$ denotes the p.d.f of a Gaussian distribution
with mean vector equal to $\mathbf{m}$ and covariance matrix equal to $\bm{\Sigma}$.

Next, the r.h.s. of (\ref{eq:posterior}) is approximated in \cite{herandezlobato2015} via expectation propagation (EP) 
to obtain $q$. For this, each non-Gaussian factor $\phi_i$ is replaced by a corresponding un-normalized Gaussian approximate 
factor $\tilde{\phi}_i$. That is,
$\phi_i(\overline{\mathbf{f}}) = \Phi\left( y_i m_i / \sqrt{s_i + 1}\right)  \approx 
\tilde{\phi}_i(\overline{\mathbf{f}}) = \tilde{s}_i \exp \{ - \tilde{\nu}_i  
	\overline{\mathbf{f}}^\text{T} \bm{\upsilon}_i\bm{\upsilon}_i^\text{T} \overline{\mathbf{f}} / 2  +  
	\tilde{\mu}_i \overline{\mathbf{f}}^\text{T} \bm{\upsilon}_i\}$,
where $\bm{\upsilon}_i=\mathbf{K}_{\overline{\mathbf{f}}\overline{\mathbf{f}}}^{-1} \mathbf{K}_{\overline{\mathbf{f}}f_i}$
is a $m$ dimensional vector, and $\tilde{s}_i$, $\tilde{\nu}_i$ and $\tilde{\mu}_i$ are parameters estimated by EP
so that $\phi_i$ is similar to $\tilde{\phi}_i$ in regions of high posterior probability as estimated by 
$q^{\setminus i}\propto q/\tilde{\phi}_i$. 
Namely, $\tilde{\phi}_i = \text{arg min}\,\, \text{KL}(\phi_i q^{\setminus i}||\tilde{\phi}_i q^{\setminus i})$, where
$\text{KL}$ is the Kullback Leibler divergence.  
We note that each $\tilde{\phi}_i$ has a one-rank precision matrix and hence only $\mathcal{O}(m)$ 
parameters need to be stored per each $\tilde{\phi}_i$.
The posterior approximation $q$ is obtained by replacing in the r.h.s. of (\ref{eq:posterior}) each exact 
factor $\phi_i$ with the corresponding $\tilde{\phi}_i$. Namely, 
$q(\overline{\mathbf{f}})=\prod_{i=1}^n \tilde{\phi}_i(\overline{\mathbf{f}}) p(\overline{\mathbf{f}}|\overline{\mathbf{X}}) / Z_q$,
where $Z_q$ is a constant that approximates $p(\mathbf{y}|\overline{\mathbf{X}})$, which can be maximized for finding
good hyper-parameters via type-II maximum likelihood \cite{rasmussen2005book}.
Finally, since all factors in $q$ are Gaussian, $q$ is a multivariate Gaussian.

In order for Gaussian process classification to work well, hyper-parameters and inducing points 
must be learned from the data. Previously, this was infeasible on big datasets using EP.
In \cite{herandezlobato2015} the gradient of $\log Z_q$ w.r.t $\xi_j$ 
(\emph{i.e.}, a parameter of the covariance function $k$ or a component 
of $\overline{\mathbf{X}}$) is:
\vspace{-.15cm}
\begin{align}
\frac{\partial \log Z_q}{\partial \xi_j} &= 
	\bm{\eta}^\text{T} \frac{\partial \bm{\theta}_\text{prior}}{\partial \xi_j} - 
	\bm{\eta}_\text{prior}^\text{T} \frac{\partial \bm{\theta}_\text{prior}}{\partial \xi_j} + 
	\sum_{i=1}^n \frac{\partial \log Z_i}{\partial \xi_j}
\,,
\label{eq:gradient}
\end{align} \par
\vspace{-.4cm}
where $\bm{\eta}$ and $\bm{\eta}_\text{prior}$ are the expected sufficient statistics under 
$q$ and $p(\overline{\mathbf{f}}|\overline{\mathbf{X}})$, respectively, $\bm{\theta}_\text{prior}$
are the natural parameters of $p(\overline{\mathbf{f}}|\overline{\mathbf{X}})$, and $Z_i$ is the normalization 
constant of $\phi_i q^{\setminus i}$. We note that 
(\ref{eq:gradient}) has a sum across the data. This enables using stochastic gradient ascent for hyper-parameter learning.

A batch iteration of EP updates in parallel each $\tilde{\phi}_i$. After this, $q$ is recomputed and the 
gradients of $\log Z_q$ with respect to each hyper-parameter are used to update the 
model hyper-parameters. The EP algorithm in \cite{herandezlobato2015} can also process data using minibatches of size $s \ll n$. 
In this case, the update of the hyper-parameters and the reconstruction of $q$ is done  
after processing each minibatch. The update of each $\tilde{\phi}_i$ corresponding to the data contained in the 
minibatch is also done in parallel. When computing the gradient of the hyper-parameters, the sum in the r.h.s. of 
(\ref{eq:gradient}) is replaced by a stochastic approximation, \emph{i.e.}, 
$n / s \sum_{i \in \mathcal{M}} \partial \log Z_i / \partial \xi_j$, 
with $\mathcal{M}$ the set of indices of the instances of the current minibatch. When using minibatches and 
stochastic gradients the training cost is $\mathcal{O}(m^3)$.

\vspace{-.3cm}

\section{Stochastic expectation propagation for training the model}

\vspace{-.2cm}

The method described in the previous section has the disadvantage that it requires to store in memory $m+1$ 
parameters for each approximate factor $\tilde{\phi}_i$. This leads to a memory cost that scales like 
$\mathcal{O}(nm)$. Thus, in very big datasets where $n$ is of the order of several millions, and in complex models where 
the number of inducing points $m$ may be in the hundreds, this cost can lead to memory problems. To alleviate this, 
we consider training via stochastic expectation propagation (SEP) as an alternative to expectation propagation \cite{Yingzhen2015}. 
SEP reduces the memory requirements by a factor of $n$.

\begin{wrapfigure}{R}{0.5\textwidth}
\vspace{-.4cm}
\begin{tabular}{ll}
\hline
\multicolumn{2}{l}{{\bf Algorithm 1:} Parallel {\bf EP} - Batch Mode} \\
\hline
{\small 1:}   & {\small For each approximate factor $\tilde{\phi}_i$ to update:} \\ 
{\small 1.1:} & \hspace{.2cm} {\small Compute cavity:} 
                {\small $q^{\setminus i}(\overline{\mathbf{f}}) \propto q(\overline{\mathbf{f}}) / \tilde{\phi}_i(\overline{\mathbf{f}})$} \\ 
{\small 1.2:} & \hspace{.2cm} {\small Update $\tilde{\phi}_i$:} 
                 $\tilde{\phi_i} = \text{{\bf proj}}(\phi_i)$\\ 
{\small 2:} & {\small Reconstruct $q$:} 
                {\small $q(\overline{\mathbf{f}}) \propto \prod_{i=1}^n 
		\tilde{\phi}_i(\overline{\mathbf{f}}) p(\overline{\mathbf{f}}|\overline{\mathbf{X}}) $} \\ 
\hline
\end{tabular} 

\vspace{.1cm}

\begin{tabular}{ll}
\hline
\multicolumn{2}{l}{{\bf Algorithm 2:} Parallel {\bf SEP} - Batch Mode} \\
\hline
{\small 1:}   & {\small Set the new global factor $\tilde{\phi}_\text{new}$ to be uniform.} \\ 
{\small 2:}   & {\small For each exact factor $\phi_i$ to incorporate:} \\ 
{\small 2.1:} & \hspace{.2cm} {\small Compute cavity:} 
                {\small $q^{\setminus i}(\overline{\mathbf{f}}) \propto q(\overline{\mathbf{f}}) / 
		\tilde{\phi}(\overline{\mathbf{f}})^\frac{1}{n}$} \\ 
{\small 2.2:} & \hspace{.2cm} {\small Find $\tilde{\phi}_i$:} 
                 $\tilde{\phi_i} = \text{{\bf proj}}(\phi_i)$\\ 
{\small 2.3:} & \hspace{.2cm} {\small Accumulate $\tilde{\phi}_i$:} 
                 $\tilde{\phi}_\text{new}(\overline{\mathbf{f}}) = \tilde{\phi}_\text{new}(\overline{\mathbf{f}})
		\tilde{\phi}_i(\overline{\mathbf{f}})$ \\ 
{\small 3:} & {\small Reconstruct $q$:} 
                {\small $q(\overline{\mathbf{f}}) \propto 
		\tilde{\phi}_\text{new}(\overline{\mathbf{f}}) p(\overline{\mathbf{f}}|\overline{\mathbf{X}}) $} \\ 
\hline
\end{tabular}

\vspace{.1cm}

\begin{tabular}{ll}
\hline
\multicolumn{2}{l}{{\bf Algorithm 3:} Parallel {\bf ADF} - Batch Mode} \\
\hline
{\small 1:}   & {\small Set $q$ to the prior. For each $\phi_i$ to process:} \\ 
{\small 1.1:} & \hspace{.2cm} {\small Compute cavity:} 
                {\small $q^{\setminus i}(\overline{\mathbf{f}}) = q(\overline{\mathbf{f}})$} \\ 
{\small 1.2:} & \hspace{.2cm} {\small Update $\tilde{\phi}_i$:} 
                 $\tilde{\phi_i} = \text{{\bf proj}}(\phi_i)$\\ 
{\small 2:} & {\small Update $q$:}
                {\small $q(\overline{\mathbf{f}}) \propto \prod_{i=1}^n \tilde{\phi}_i(\overline{\mathbf{f}}) q(\overline{\mathbf{f}}) $} \\ 
\hline
\end{tabular} 

\caption{Comparison among EP, SEP and ADF in the model from \cite{herandezlobato2015}. Training is done 
in batch mode. The projection step refers to KL minimization:  $\text{{\bf proj}}(\phi_i)={\text{arg min}}_{\tilde{\phi}_i}\,\, 
\text{KL}(\phi_i q^{\setminus i}||\tilde{\phi}_i q^{\setminus i})$. }
\label{fig:fig_EP_vs_SEP}
\vspace{-1.25cm}
\end{wrapfigure}

In SEP the likelihood of the model is approximated by a single global Gaussian factor 
$\tilde{\phi}$, instead of a product of $n$ Gaussian factors $\tilde{\phi}_i$. 
The idea is that the natural parameters $\bm{\theta}$ of $\tilde{\phi}$ approximate the sum of the natural 
parameters $\sum_{i=1}^n \bm{\theta}_i$ of the EP approximate factors $\tilde{\phi}_i$. This approximation 
reduces by a factor of $n$ the memory requirements because only the natural parameters $\bm{\theta}$ 
of $\tilde{\phi}$ need to be stored in memory, and the size of $\bm{\theta}$ is dominated by the precision matrix 
of $\tilde{\phi}$, which scales like $\mathcal{O}(m^2)$.

When SEP is used instead of EP for finding $q$ some things change. In particular, the computation of the 
cavity distribution $q^{\setminus i}\propto q / \tilde{\phi}_i$ is now replaced by 
$q^{\setminus i}\propto q / \tilde{\phi}^{1 / n}$, $\forall i$. Furthermore, in the case of the batch learning 
method described in the previous section, the corresponding approximate factor $\tilde{\phi}_i$ for
each instance is computed as $\tilde{\phi}_i = \text{arg min}\,\, \text{KL}(\phi_i q^{\setminus i}||\tilde{\phi}_i q^{\setminus i})$
to then set $\tilde{\phi} = \prod_{i=1}^n \tilde{\phi}_i$. This is equivalent to adding natural parameters,
\emph{i.e.}, $\bm{\theta} = \sum_{i=1}^n \bm{\theta}_i$. In the case of minibatch training with minibatches of size 
$s\ll n$ the update is slightly different to account for the fact that we have only processed a small amount of the 
total data. In this case, $\bm{\theta}_\text{new} = \bm{\theta}_\text{old} \times (n - s) / n  + \sum_{i \in \mathcal{M}} \bm{\theta}_i$,
where $\mathcal{M}$ is a set with the indices of the instances contained in the current minibatch.
Finally, in SEP the computation of the gradients for updating the hyper-parameters is done exactly as in EP.
Figure \ref{fig:fig_EP_vs_SEP} compares among EP, SEP and ADF \cite{maybeck82} when used 
to update $q$. In the figure training is done in batch mode and the update of the hyper-parameters has been 
omitted since it is exactly the same in either EP, SEP or ADF. In ADF the cavity distribution 
$q^{\setminus i}$ is simply the posterior approximation $q$, and when $q$ is recomputed, the natural parameters 
of the approximate factors are simply added to the natural parameters of $q$. ADF is a simple baseline in which 
each data point is \emph{seen} by the model several times and hence it underestimates variance \cite{Yingzhen2015}.

\vspace{-.4cm}

\section{Experiments}

\vspace{-.3cm}

We evaluate the performance of the model described before when trained using EP, SEP and ADF. 

{\bf Performance on datasets from the UCI repository:}
First, we consider 7 datasets from the UCI repository. The experimental protocol
followed is the same as the one described in \cite{herandezlobato2015}. In these experiments we consider
a different number of inducing points $m$. Namely, $15\%$, $25\%$ and $50\%$ of the total training instances and
the training of all methods is done in batch mode for 250 iterations. Table \ref{tab:ll_uci} shows the average 
negative test log likelihood  of each method (the lower the better) on the test set. The best method 
has been highlighted in boldface. We note that SEP obtains similar and sometimes even better results than EP. By contrast, 
ADF performs worse, probably because it underestimating the posterior variance. In terms of the average training 
time all methods are equal.

\begin{table}[htb]
\begin{center}
\caption{\small Average negative test log likelihood for each method and average training time in seconds. }
\label{tab:ll_uci}
{\small
\begin{tabular}{@{\hspace{.3mm}}l@{\hspace{1mm}}|@{\hspace{.3mm}}c@{{\tiny $\pm$}}c@{\hspace{0.3mm}}c@{{\tiny $\pm$}}c@{\hspace{0.3mm}}c@{{\tiny $\pm$}}c@{\hspace{.3mm}}|@{\hspace{.3mm}}c@{{\tiny $\pm$}}c@{\hspace{0.3mm}}c@{{\tiny $\pm$}}c@{\hspace{0.3mm}}c@{{\tiny $\pm$}}c@{\hspace{.3mm}}|@{\hspace{.3mm}}c@{{\tiny $\pm$}}c@{\hspace{0.3mm}}c@{{\tiny $\pm$}}c@{\hspace{0.3mm}}c@{{\tiny $\pm$}}c@{\hspace{.3mm}}}
\hline
& \multicolumn{6}{c}{$m=15\%$} 
& \multicolumn{6}{|c}{$m=25\%$} 
& \multicolumn{6}{|c}{$m=50\%$} \\
\hline
\bf{Problem}  & 
	\multicolumn{2}{c}{\footnotesize \bf ADF} & \multicolumn{2}{c}{\bf EP} & \multicolumn{2}{c|}{\bf SEP}  &
	\multicolumn{2}{c}{\footnotesize \bf ADF} & \multicolumn{2}{c}{\bf EP} & \multicolumn{2}{c|}{\bf SEP}  &
	\multicolumn{2}{c}{\footnotesize \bf ADF} & \multicolumn{2}{c}{\bf EP} & \multicolumn{2}{c}{\bf SEP}  \\
\hline        
Australian    &   .70    &   .07    &   .69    &   .07    &  \bf{ .63 }   &  \bf{ .05 }    &      .70    &  .08    &   .67    &   .07    &  \bf{ .63 }   &  \bf{ .05 }   &    .67    &   .06    &   .64    &   .05    &  \bf{ .63 }   &  \bf{ .05 }   \\ 
Breast        &   .12    &   .06    &   .11    &   .05    &  \bf{ .11 }   &  \bf{ .05 }   &       .12    &  .05    &   .11    &   .05    &  \bf{ .11 }   &  \bf{ .05 }   &    .12    &   .05    &  \bf{ .11 }  &  \bf{ .05 }  &   .11    &   .06        \\ 
Crabs         &   .08    &   .06    &   .06    &   .06    &  \bf{ .06 }   &  \bf{ .07 }   &       .09    &  .06    &   .06    &   .06    &  \bf{ .06 }   &  \bf{ .07 }   &  .08    &   .06    &   .06    &   .06    &  \bf{ .06 }   &  \bf{ .07 }  \\
Heart         &   .45    &   .18    &   .40    &   .13    &  \bf{ .39 }   &  \bf{ .11 }    &      .48    &  .18    &   .41    &   .12    &  \bf{ .40 }   &  \bf{ .11 }   &    .46    &   .17    &   .41    &   .11    &  \bf{ .40 }   &  \bf{ .12 }         \\
Ionosphere    &   .29    &   .18    &  \bf{ .26 }   &  \bf{ .19 }   &   .28    &   .16    &       .30    &  .17    &   .27    &   .20    &  \bf{ .27 }   &  \bf{ .17 }   &    .33    &   .19    &   .27    &   .19    &  \bf{ .27 }   &  \bf{ .17 }      \\
Pima          &   .52    &   .07    &   .52    &   .07    &  \bf{ .49 }   &  \bf{ .05 }   &       .58    &  .10    &   .51    &   .06    &  \bf{ .49 }   &  \bf{ .05 }   &    .62    &   .09    &   .50    &   .05    &  \bf{ .49 }   &  \bf{ .05 }    \\
Sonar         &   .40    &   .15    &  \bf{ .33 }   &  \bf{ .10 }   &   .35    &   .11      &     .46    &  .32    &  \bf{ .32 }   &  \bf{ .10 }   &   .35    &   .12    &    .46    &   .24    &  \bf{ .29 }   &  \bf{ .09 }   &   .33    &   .12           \\
\hline        
{\bf Avg. Time } & 18.2 &   0.3 &   19.3 &   0.5 &   18.8 &   0.1 & 38.5 &   0.5 &   38.9 &   0.6 &   40.2 &   0.2 &
  145 &   4.0 &   136 &   3.0 &   149 &   1.0 \\
\hline
\end{tabular}
}
\end{center}
\vspace{-.2cm}
\end{table}

\begin{figure}[htb]
\vspace{-.35cm}
\begin{center}
\begin{tabular}{cc}
\includegraphics[width = 0.475 \textwidth]{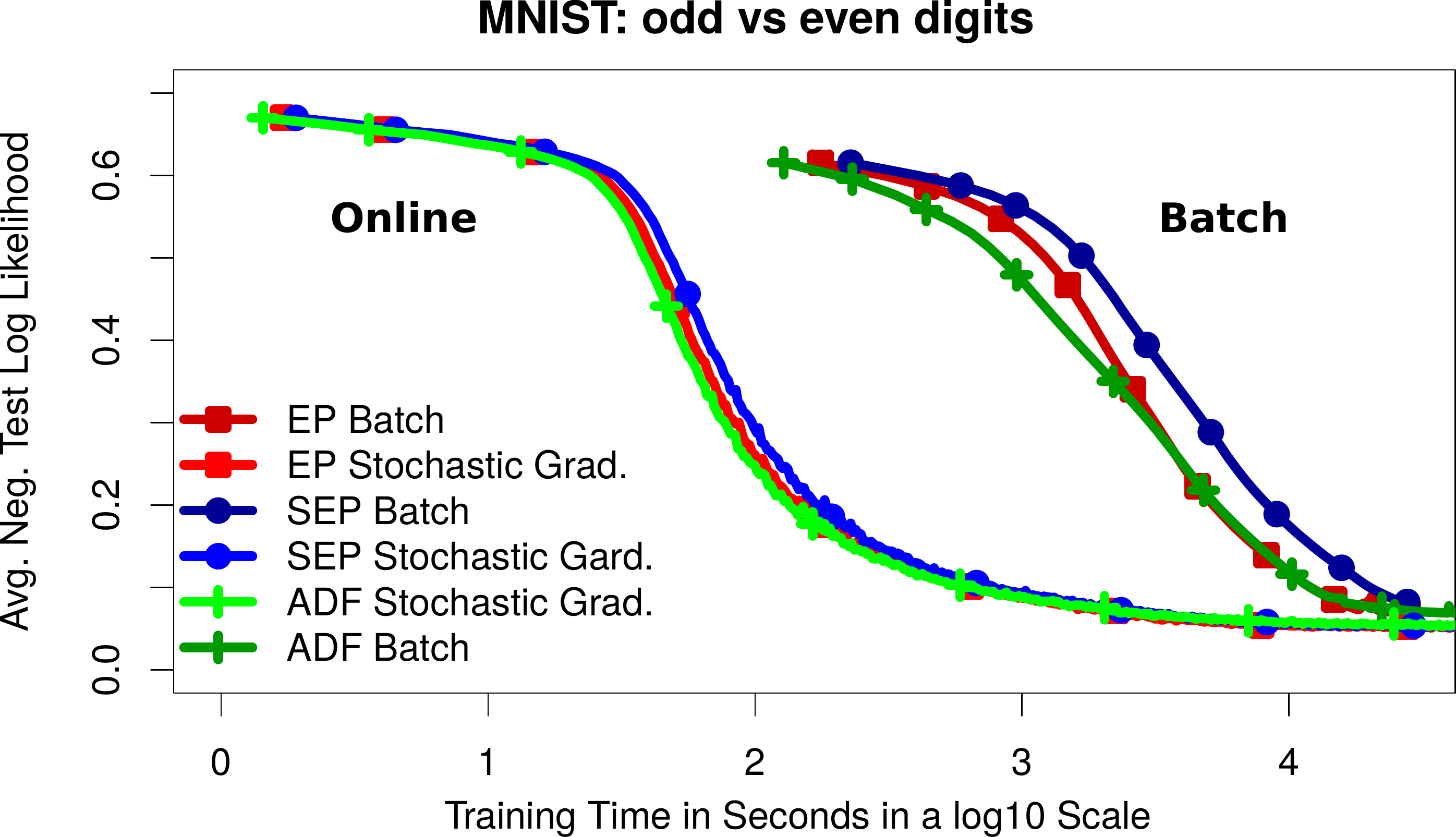}  &
\includegraphics[width = 0.475 \textwidth]{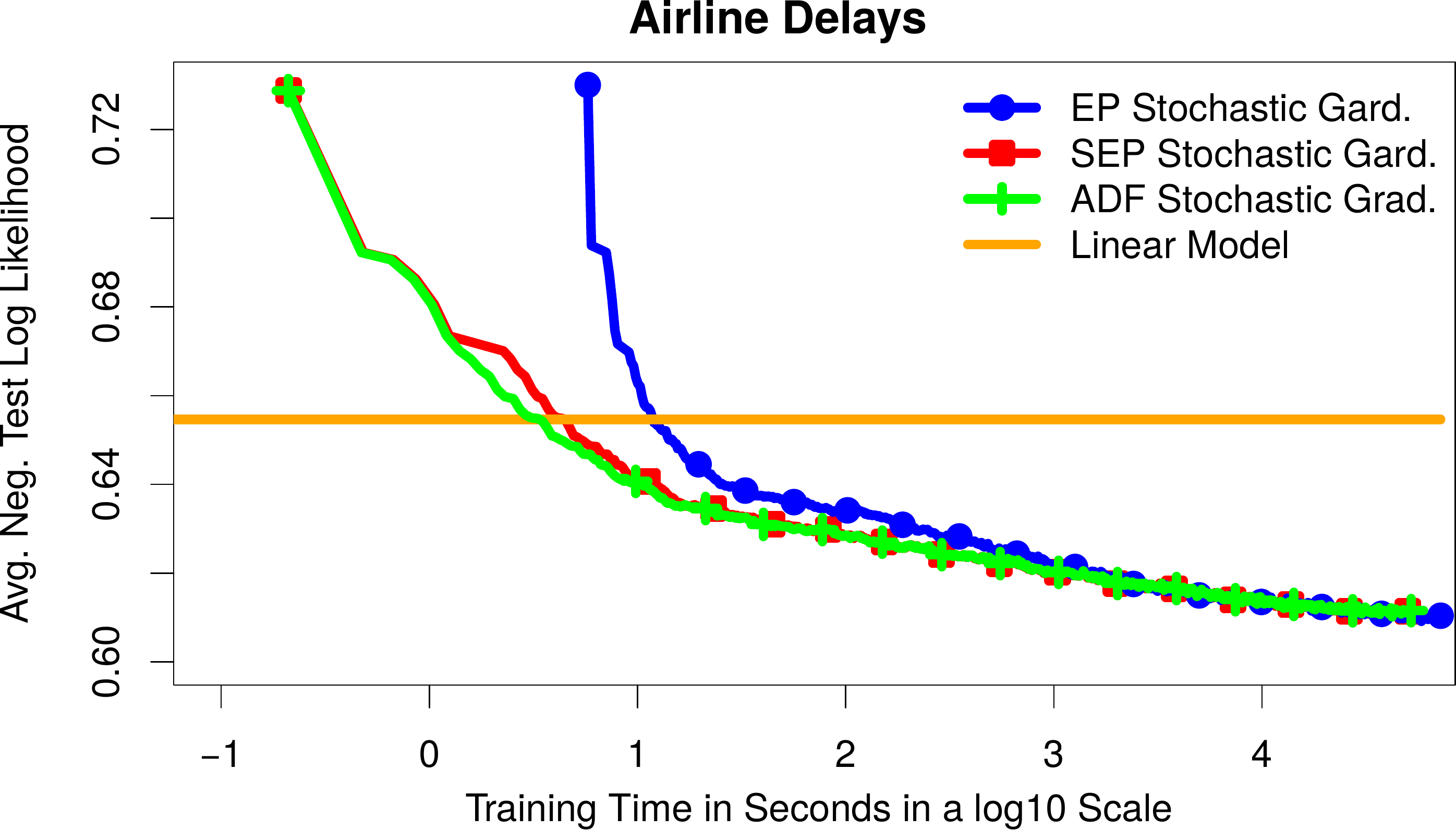}   
\end{tabular}
\end{center}
\vspace{-.55cm}
\caption{{\small (top) Avg. negative test log likelihood for EP, SEP and ADF as a function
of time. We show results when using a minibatch size 
$200$ for training (stochastic) and when using all data instances at once (batch).
The performance of a linear logistic regression classifier is also shown for the airline dataset. 
Best seen in color. }}
\label{fig:stochastic}
\vspace{-.6cm}
\end{figure}

{\bf Performance on big datasets:} We carry out experiments when the model is trained using minibatches. 
We follow \cite{herandezlobato2015} and consider the MNIST dataset, which has 70,000 instances, and the 
airline delays dataset, which has 2,127,068 data instances (see \cite{herandezlobato2015} for more details). 
In both cases the test set has 10,000 instances. Training is done using minibatches of size 200, which is 
equal to the number of inducing points $m$. In the case of the MNIST dataset we also report results for 
batch training (in the airline dataset batch training is infeasible). Figure \ref{fig:stochastic} shows the 
avg. negative log likelihood obtained on the test set as a function of training time. In the MNIST dataset 
training using minibatches is much more efficient. Furthermore, in both datasets SEP performs very similar to EP. 
However, in these experiments ADF provides equivalent results to both SEP and EP. Furthermore, in the airline dataset both SEP and 
ADF provide better results than EP at the early iterations, and improve a simple linear model 
after just a few seconds. The reason is that, unlike EP, SEP and ADF do not initialize the approximate 
factors to be uniform, which has a significant cost in this dataset. 

\begin{wrapfigure}{r}{0.5\textwidth}
\vspace{-0.3cm}
\includegraphics[width = 0.5 \textwidth]{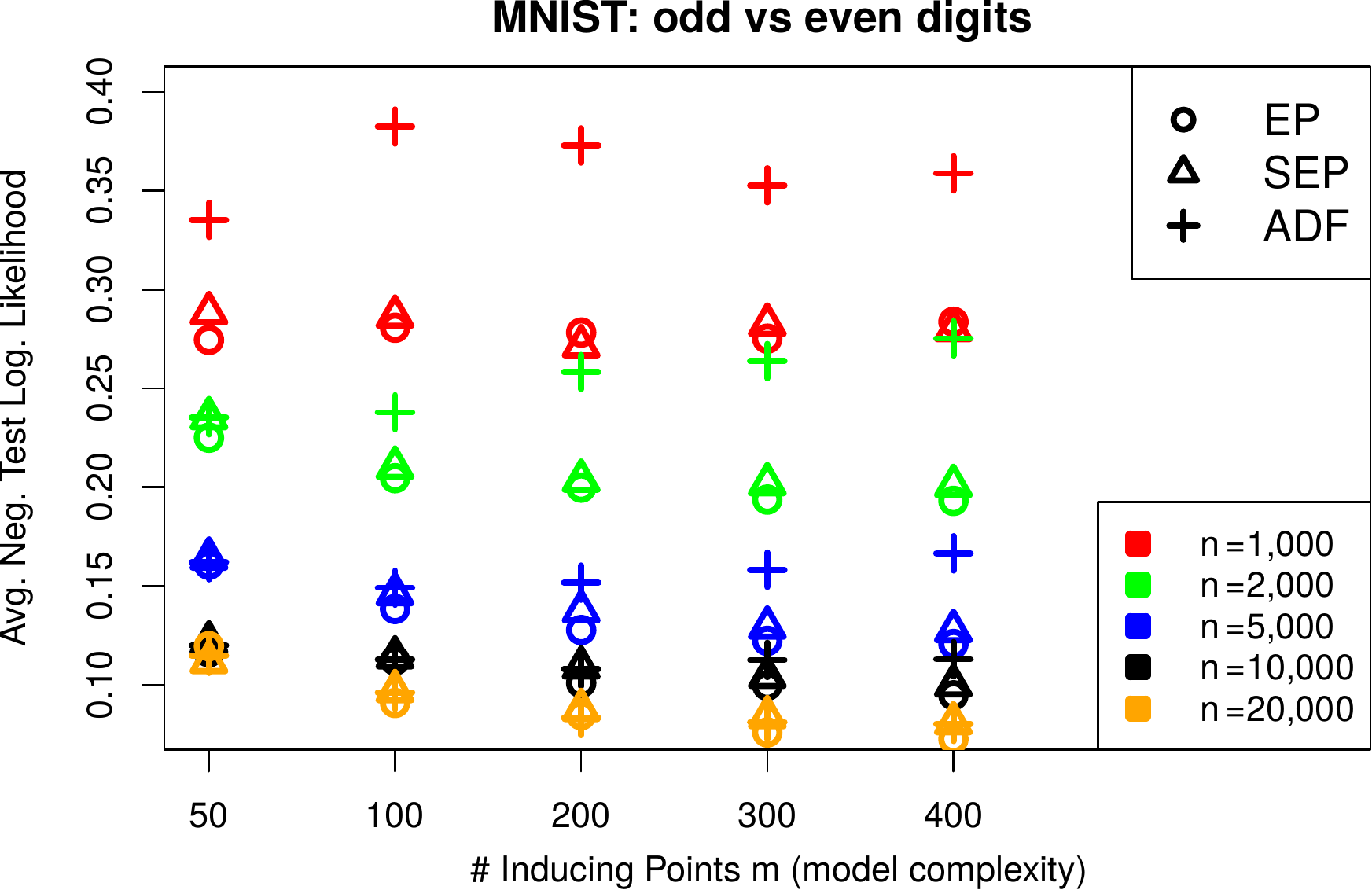} 
\vspace{-0.75cm}
\caption{{\small Performance of each method (ADF, SEP and EP) on the MNIST dataset for
increasing training set sizes $n$ and model complexity (\# inducing points $m$).}}
\label{fig:n_vsm}
\vspace{-0.3cm}
\end{wrapfigure}

$\text{{\bf Dataset size and model complexity:}}$
The results obtained in the large datasets contradict the results obtained in 
the UCI datasets in the sense that ADF performs similar to EP. We believe the 
reason for this is that ADF may perform similar to EP only when the model is simple 
(small $m$) and/or when the number of training instances is very large (large $n$). 
To check that this is the case, we repeat the experiments with the MNIST dataset
with an increasing number of training instances  $n$ (from $1,000$ to $20,000$) 
and with an increasing number of inducing points $m$ (from $50$ to $400$).
The results obtained are shown in Figure \ref{fig:n_vsm}, which confirms that
ADF only performs similar to EP in the scenario described. By contrast, SEP 
seems to always perform similar to EP. Finally, increasing the model complexity ($m$) 
seems beneficial.

\vspace{-.4cm}

\section{Conclusions}

\vspace{-.3cm}

Stochastic expectation propagation (SEP) 
\cite{Yingzhen2015} can reduce the memory cost of the method recently proposed in \cite{herandezlobato2015} to 
address large scale Gaussian process classification. Such a method 
uses expectation propagation (EP) for training, which stores $\mathcal{O}(nm)$ parameters in memory, 
where $m \ll n$ is some small constant and $n$ is the training set size. This cost may be too expensive in the case of 
very large datasets or complex models. SEP reduces the storage resources needed by a factor of $n$, leading to a 
memory cost that is $\mathcal{O}(m^2)$. Furthermore, several experiments show that SEP provides similar 
performance results to those of EP. A simple baseline known as ADF may also provide similar results to SEP, 
but only when the number of instances is very large and/or the chosen model is very simple.
Finally, we note that applying Bayesian learning methods at scale makes most sense with 
large models, and this is precisely the aim of the method described in this extended abstract. 

{\bf Acknowledgments:}
YL thanks the Schlumberger Foundation for her Faculty for the Future PhD fellowship. 
JMHL acknowledges support from the Rafael del Pino Foundation. 
RET thanks EPSRC grant \#s EP/G050821/1 and EP/L000776/1. 
TB thanks Google for funding his European Doctoral Fellowship.
DHL and JMHL acknowledge support from Plan Nacional I+D+i, Grant TIN2013-42351-P, 
and from Comunidad Aut\'onoma de Madrid, Grant S2013/ICE-2845 CASI-CAM-CM.
DHL is grateful for using the computational resources of 
\emph{Centro de Computaci\'on Cient\'ifica} at Universidad Aut\'onoma de Madrid.

%\bibliography{references}

\bibliographystyle{unsrt}

\end{document}